\documentclass[aps, showkeys]{revtex4-2}
\usepackage{graphicx, xcolor, amssymb, amsmath, amsfonts}\usepackage[colorlinks=true, linkcolor=blue, citecolor=blue, urlcolor=blue]{hyperref}
\usepackage{url, soul, ulem, textgreek, bm, lineno}
\usepackage[draft]{changes}
\definechangesauthor[name = Kyoung-Min, color = blue]{KM}
\definechangesauthor[name = Tilen, color = red]{T}

\begin{document}

\title{Neural Scaling Laws for Deep Regression}

\author{Tilen \v{C}ade\v{z}}
\affiliation{Asia Pacific Center for Theoretical Physics, Pohang, Gyeongbuk, 37673, Republic of Korea}

\author{Kyoung-Min Kim}
\email{kyoungmin.kim@apctp.org}
\affiliation{Asia Pacific Center for Theoretical Physics, Pohang, Gyeongbuk, 37673, Republic of Korea}
\affiliation{Department of Physics, Pohang University of Science and Technology, Pohang, Gyeongbuk 37673, Korea}

\begin{abstract}
    Neural scaling laws—power-law relationships between generalization errors and characteristics of deep learning models—are vital tools for developing reliable models while managing limited resources. Although the success of large language models highlights the importance of these laws, their application to deep regression models remains largely unexplored. Here, we empirically investigate neural scaling laws in deep regression using a parameter estimation model for twisted van der Waals magnets. We observe power-law relationships between the loss and both training dataset size and model capacity across a wide range of values, employing various architectures—including fully connected networks, residual networks, and vision transformers. Furthermore, the scaling exponents governing these relationships range from 1 to 2, with specific values depending on the regressed parameters and model details. The consistent scaling behaviors and their large scaling exponents suggest that the performance of deep regression models can improve substantially with increasing data size.
\end{abstract}

\maketitle

\section{Introduction}

\begin{figure}[h!]
    \centering
   \includegraphics[width=0.6\textwidth]{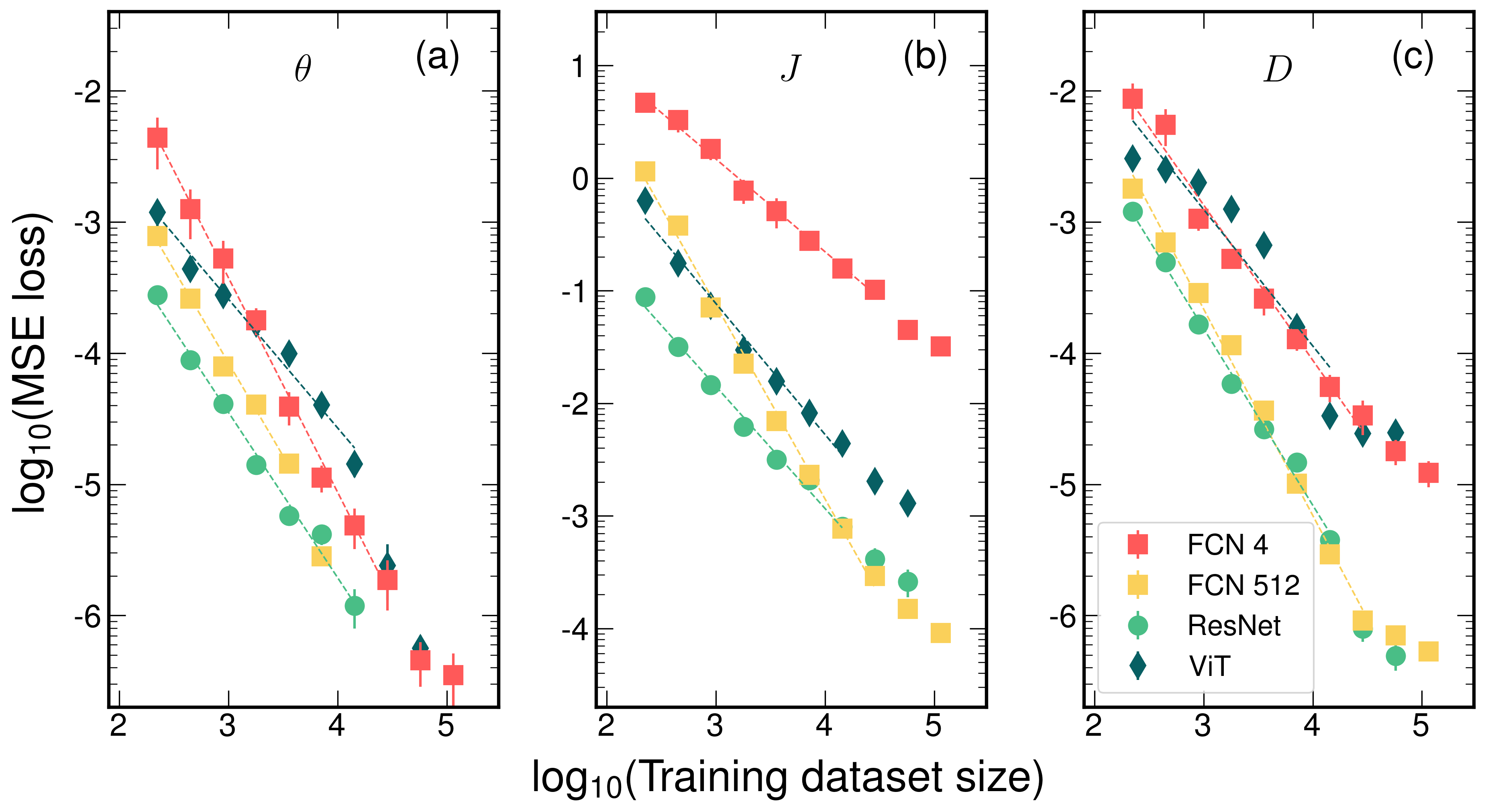}
    \caption{
    \textbf{Neural scaling laws in a parameter estimation model.} Panels (a)–(c) show the geometric mean of the mean squared error (MSE) test loss in the prediction of three magnetic parameters (\(\theta\), \(J\), \(D\)), respectively, as a function of training dataset size. Each marker corresponds to a different architecture: fully connected networks (FCNs), residual network (ResNet-18), and vision transformer (ViT). The two FCNs, indicated by red and yellow squares, have different numbers of neurons per hidden layer (\(n_n=4\) and \(n_n=512\), respectively) with the same number of hidden layers (\(n_l=3\)). In each panel, dashed lines represent power-law fits of the form \(\epsilon \sim N_D^{-\alpha_D}\), where \(\epsilon\) is the loss, \(N_D\) is the training dataset size, and \(\alpha_D\) is the scaling exponent. All data are plotted on a log-log scale, with dataset sizes ranging from 224 to 114688. 
    }
    \label{fig1}
\end{figure}

Deep learning has established a new data-driven programming paradigm, where data determines the algorithms used to solve specific problems. This approach often provides substantial advantages over traditional logic-driven programming, particularly in addressing intricate scientific and real-world problems, where the required logic of algorithms is unknown or too complex. Generally, the performance of deep learning models depends on the quantity and quality of the data used to train deep neural networks. Researchers have identified various techniques for enhancing model performance, either by utilizing various types of regularization and batch normalization from the model side, or by pre-processing a given dataset \cite{8686063} or through data augmentation \cite{perez2017effectiveness, Steiner2022HowViT} from the data side. The latter potential can be systematically examined using ``generalization error scaling laws," which describe the power-law relationship between generalization errors and dataset size, model size and the computing time used for the training.

Several early~\cite{AMARI1993161, 6796972} as well as more recent~\cite{maloney2022solvable,Defilippis2023DimensionFree,Paquette2024phases, bahri2024explaining} theoretical studies sought to derive generalization error scaling laws by solving statistical models, while recent research has focused on empirical investigations in the context of classification models \cite{BAILLY2022106504, hestness2017deeplearningscalingpredictable, rosenfeld2020constructive, henighan2020scaling, mühlenstädt2024dataneed2predicting, meir2020powerlaw, Figueroa2012, cho2016dataneededtrainmedical, 8237359, Figueroa2012, JMLR:v23:20-1111, NEURIPS2022_8c22e5e9} and language models \cite{kaplan2020scalinglawsneurallanguage, NEURIPS2022_8c22e5e9, hoffmann2022trainingcomputeoptimallargelanguage, NEURIPS2022_8c22e5e9, JMLR:v23:20-1111, kaplan2020scalinglawsneurallanguage, hutter2021learningcurvetheory, NEURIPS2022_8c22e5e9}. In particular, the exploration of the scaling laws in large language models found that the sizes of datasets and models are crucial for model performance, while other features such as network structure have secondary impacts \cite{kaplan2020scalinglawsneurallanguage, bahri2024explaining}. These insights have significantly contributed to the recent success of large language models. In contrast to substantial advancements in classification and language models, the scaling laws for deep regression models remain largely understudied, despite recent emerging research \cite{lin2024scalinglawslinearregression}. Gaining a deeper understanding of these laws could provide valuable insights for developing deep regression models with predictable performance. Such insights are particularly crucial in scientific disciplines where computational and data resources are often limited and optimizing resource usage is important. 

\begin{figure}[t!]
    \centering
   \includegraphics[width=.8\textwidth]{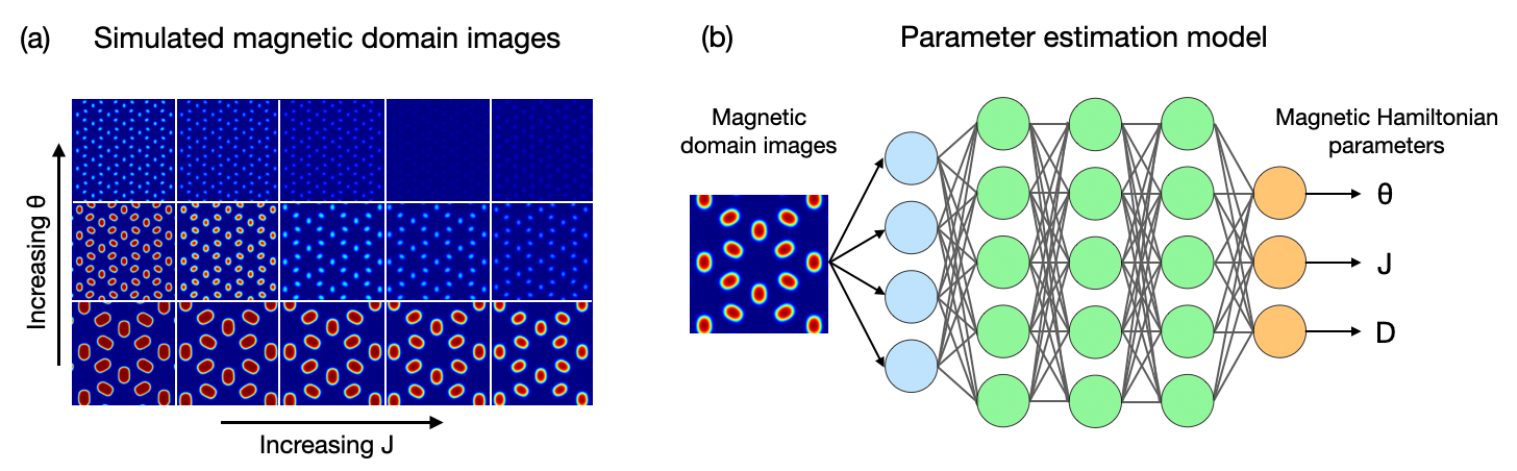}
    \caption{
    \textbf{Dataset and regression model used in the scaling analysis of generalization error.} (a) Dataset: simulated magnetic domain images generated via atomistic spin simulations of twisted bilayer CrI\textsubscript{3}. Red (blue) indicates \(+1\) \((-1)\) of the local out-of-plane normalized magnetization. Only images of the top layer are shown. (b) Regression model: the parameter estimation model predicts the magnetic Hamiltonian parameters \((\theta, J, D)\) for twisted bilayer CrI\textsubscript{3} from the input magnetic domain images.
    }
    \label{fig:model}
\end{figure}

In this work, we empirically investigate the scaling laws of generalization error in deep regression, using a parameter estimation model that predicts magnetic Hamiltonian parameters from input magnetic domain images [Fig. \ref{fig:model}]. By training networks on datasets of varying sizes or on multiple networks with a fixed dataset, we examine how the mean squared error (MSE) test loss in the model's predictions scales with the training dataset size. We find that the test loss follows a power-law scaling behavior, expressed as \(\epsilon \sim N_D^{-\alpha_D}\), where \(\epsilon\) is the MSE test loss, \(N_D\) is the dataset size, and \(\alpha_D\) is the corresponding scaling exponent [Fig. \ref{fig1}]. Notably, the scaling behavior often spans more than three orders of magnitude in dataset size. Furthermore, the scaling exponents \(\alpha_D\) range from approximately 1 to 2—much larger than those typically observed in classification or language models, which are often below 1 \cite{BAILLY2022106504, hestness2017deeplearningscalingpredictable, rosenfeld2020constructive, henighan2020scaling, meir2020powerlaw, mühlenstädt2024dataneed2predicting, Figueroa2012, cho2016dataneededtrainmedical, 8237359, Figueroa2012, JMLR:v23:20-1111, NEURIPS2022_8c22e5e9, kaplan2020scalinglawsneurallanguage, hoffmann2022trainingcomputeoptimallargelanguage,  hutter2021learningcurvetheory}, with a rare case slightly above 1 \cite{bahri2024explaining} and a recent study~\cite{maloney2022solvable} with exponents of up to 2. This suggests that the model’s performance improves more rapidly with increased data or model complexity than typically expected. Additionally, increasing model complexity significantly reduces the MSE test loss, showing a similar power-law scaling \(\epsilon \sim N_M^{-\alpha_M}\) with the model size \(N_M\) and the scaling exponent \(\alpha_M\) [Fig. \ref{fig6}]. These findings demonstrate that deep regression models can achieve substantial improvements in accuracy as data and model sizes grow.

\section{Result}

\subsection{Neural network realizations}
\label{subsection:statistics}

\begin{figure}[t!]
    \centering
    \includegraphics[width=.9\textwidth]{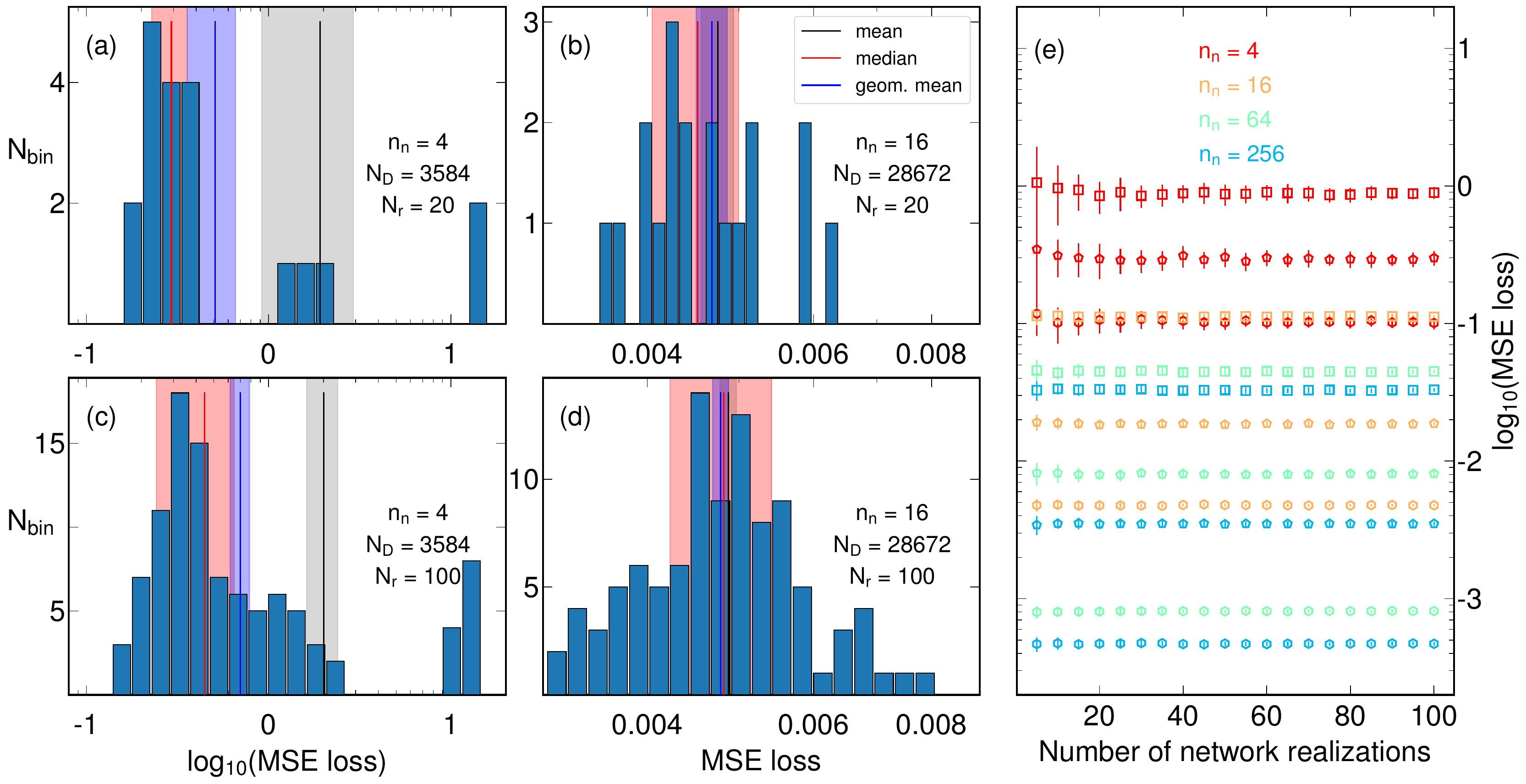}
    \caption{
    \textbf{Statistics of MSE test loss across different realizations of FCNs.} (a)–(d) Distributions of test loss across multiple network realizations \(N_r\) for the regression of \(J\), with bars indicating the number of networks \(N_{\mathrm{bin}}\) within each loss range. Each panel corresponds to a specific combination of the number of neurons per hidden layer \(n_n\) and dataset size \(N_D\), with a fixed number of hidden layers \(n_l=3\): panels (a) and (c) show results for \(n_n=4\) and \(N_D = 3584\); panels (b) and (d) are for \(n_n=16\) and \(N_D = 28672\). Panels (a) and (b) display distributions over \(N_r = 20\) network realizations, whereas panels (c) and (d) show results with \(N_r = 100\) realizations. In each panel, black, blue, and red lines denote the arithmetic mean, geometric mean, and median of each distribution, respectively, with shaded regions representing the standard errors of the arithmetic and geometric means and the median absolute deviation, each depicted in their corresponding colors. (e) Evolution of the geometric mean with increasing number of network realizations. Results for four model sizes (\(n_n=4, 16, 64, 256\)) are shown in red, orange, green, and blue. For each size, data are presented across three testing dataset sizes \(N_D = 1792\), 7168, and 28672, marked by square, pentagon, and hexagon symbols, respectively. Error bars indicate the standard deviation of the geometric means across bootstrap samples. The bootstrap procedure involves generating 50 random subsets from the data of 100 networks, computing the geometric mean for each subset, and plotting the average of these means.
    }
    \label{fig3}
\end{figure}

Our atomistic spin simulations of twisted bilayer CrI\textsubscript{3} \cite{Kim2023, PhysRevB.108.L100401, Kim2024, Lee_2024, Kim2025} generated 162782 pairs of simulated magnetic domain images. These magnetic domain images exhibit regular yet complex variation patterns in response to changes in the underlying spin model parameters \cite{PhysRevResearch.3.013027, Zheng2022, Kim2023, PhysRevB.108.L100401, Kim2024, Kim2025}, as illustrated in Fig.~\ref{fig:model}(a). The parameter estimation model employed in this study takes local out-of-plane magnetization, represented as two-dimensional arrays, as input, and predicts the underlying magnetic Hamiltonian parameters \((\theta, J, D)\) as output [Fig.~\ref{fig:model}(b)]. Using various neural network architectures—including fully-connected networks (FCNs), ResNet-18 \cite{He2016-ResNet}, and a vision transformer (ViT) \cite{dosovitskiy2021image}—we investigate how the MSE test loss in parameter prediction scales with the training dataset size. We use MSE test loss as our primary metric to measure the network's performance throughout this work. For FCNs, we examine a range of model sizes by varying either the number of neurons per hidden layer (with a fixed number of hidden layers) or the number of hidden layers (with a fixed number of neurons in each layer) to investigate how the model size influences the scaling behavior of the test loss. The training datasets are randomly sampled, and the values of the test loss are averaged over multiple network realizations. Each realization is defined as an independently trained instance of the network, starting with a random initialization for its weights and biases. A consistent test dataset is used throughout this study. Further details on the atomistic spin simulations and deep learning methods can be found in the Methods section.

Figures~\ref{fig3}(a)–(d) illustrate the test loss distributions among various network realizations within the same network architecture and training dataset size. The test loss distributions are not uniform due to the intrinsic stochasticity of random sampling and the inherent randomness in the network optimization process~\cite{Mitchell1997}. The non-uniformity of the distribution is particularly significant in smaller models where outliers with relatively poor performance incidentally appear [Figs.~\ref{fig3}(a) and (c)]. The presence of outliers biases the arithmetic mean, rendering it an unreliable measure of average performance, especially in smaller models. In contrast, the geometric mean exhibits greater robustness to outliers and provides a more accurate representation of the test loss distribution. Accordingly, we adopt the geometric mean to evaluate the average performance across different network realizations. Using the geometric mean provides a consistent framework for assessing model performance across various network sizes.

Figure~\ref{fig3}(e) shows the geometric mean of the test loss as a function of the number of trained networks used in averaging. The geometric mean for $n_n = \{16, 64, 256\}$ is relatively constant across the number of realizations. In contrast, the geometric mean for $n_n = 4$ shows substantial fluctuations at small realization counts while plateaus once a sufficiently large number of network realizations is included. Increasing the number of realizations also decreases the variance, indicating that reliable estimates of the average performance can be achieved with a large enough sample size. In our analysis, we utilize twenty network realizations to estimate the geometric mean of the test loss, providing a robust measure across all considered architectures. Notably, larger models tend to have smaller variances, and increasing the dataset size further reduces the variance. In other words, different network realizations in the large model and dataset regimes demonstrate consistent results. This uniformity suggests that the derived scaling laws are more reliable and possess strong predictive power for real-world applications, where constructing only a limited number of models is typical, unlike the extensive stochastic averaging employed in this study.

We found that accurately capturing the scaling behavior of the test loss requires careful tuning of the learning rate during training. For smaller models, a fixed learning rate ensures consistent power-law scaling up to the largest dataset size used. However, for large models, using a fixed learning rate results in substantial fluctuations in test loss as a function of dataset size, rather than the expected power-law relationships across a large dataset size regime. In contrast, gradually decreasing the learning rate enables optimization of model performance across different dataset sizes, thereby revealing well-established scaling behavior. A key difference between the fixed and adjustable learning rate schemes lies in the scaling exponent values and the applicable scaling regimes. To examine the power-law scaling laws across the largest dataset size regime, we employ an adjustable learning rate scheme throughout this work; the detailed methodology is presented in the Methods section. Furthermore, we tested standard regularization techniques such as L2 regularization and dropout; however, these approaches did not enhance the consistency of the observed scaling behaviors. Detailed analyses are provided in the Supplementary Information.

\subsection{Scaling laws with the data set size}
\label{subsection:dataset}

\begin{figure}[t!]
    \centering
    \includegraphics[width=0.7\textwidth]{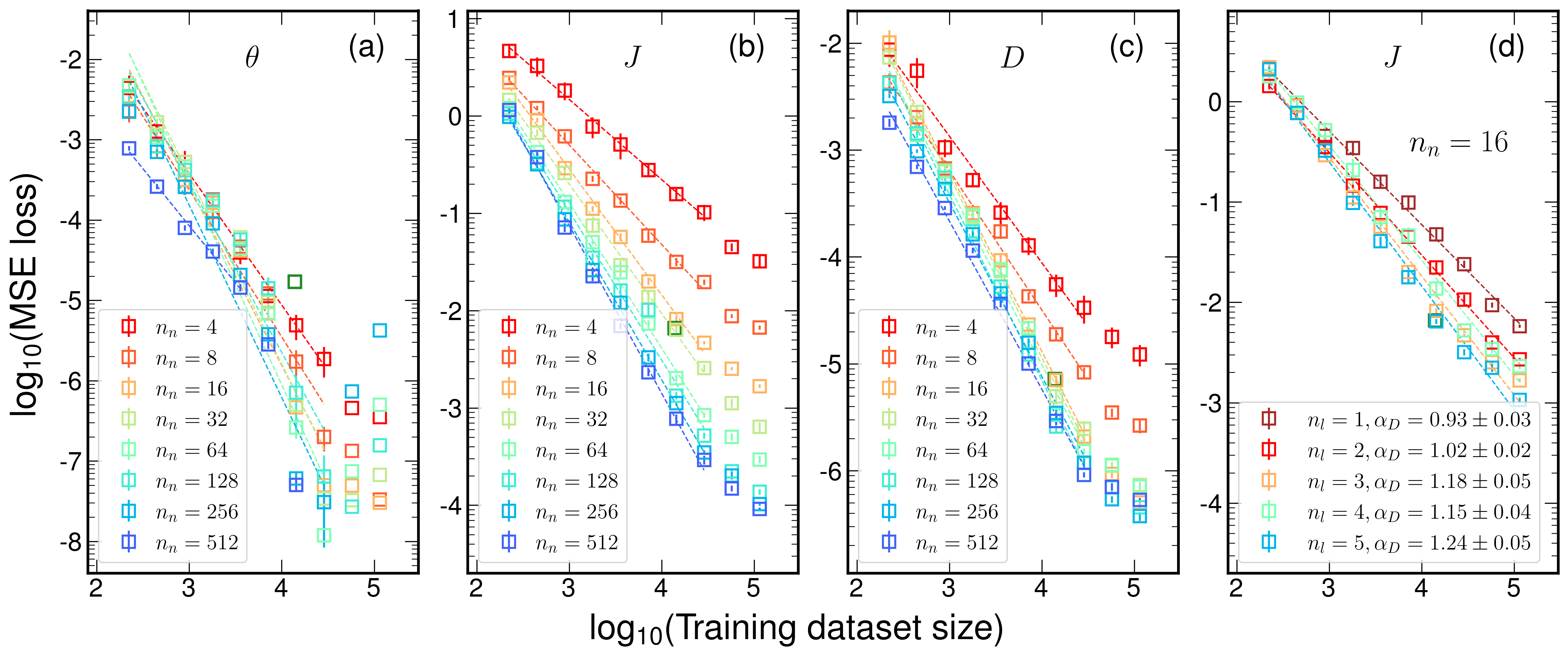}
    \caption{
    \textbf{Geometric mean of MSE test loss as a function of training dataset size.} Panels (a)–(c) show results for the regression of \(\theta\), \(J\), and \(D\), respectively. Each marker represents FCNs with varying \(n_n\) from 4 to 512 with a fixed value of \(n_l=3\). Panel (d) presents the result for the regression of \(J\), with each marker indicating different values of \(n_l\) and a fixed value of \(n_n=16\). In each panel, dashed lines are power-law fits of the form \(\epsilon \sim N_D^{-\alpha_D}\). In panel (d), the fitted values of \(\alpha_D\) are included in the legend. All data are plotted on a log-log scale, with dataset sizes ranging from 224 to 114688. The green square in panels (a)--(c) indicates the result reported in Ref.~\citenum{Lee_2024}.
    }
    \label{fig4}
\end{figure}

Figure~\ref{fig4} illustrates the variation of the test loss in predicting magnetic Hamiltonian parameters as a function of training dataset sizes across different FCN architectures. The test loss displays a monotonically decreasing trend with increasing dataset size for all regressed parameters and network architectures. These trends are well captured by a power-law scaling formula \(\epsilon \sim N_D^{-\alpha_D}\), as indicated by the dashed lines. Notably, this power-law scaling is most evident in the small to intermediate dataset regimes (up to 28,672, as shown in the plots), spanning several orders of magnitude in both test loss and dataset size. Beyond this regime, the scaling breaks down. For \(\theta\), the test loss exhibits substantial fluctuations in this regime [Fig.~\ref{fig4}(a)], likely due to the numerical precision limit of approximately \(1.9 \times 10^{-7}\) imposed by the 32-bit floating-point calculations used in our computations of regressed parameters and corresponding MSE test loss, thus values below this are considered unreliable. In contrast, for parameters \(J\) and \(D\), the loss displays slowing down from the trend [Figs.~\ref{fig4}(b) and (c)]. It is worth noting that the FCN with approximately \(3.2\times10^5\) parameters achieves a test loss comparable to that reported in Ref.~\citenum{Lee_2024} (which used an FCN with approximately \(2.1 \times 10^{7}\) parameters) for the regression of \(J\) and \(D\), while showing a substantial deviation for \(\theta\).

Figure~\ref{fig4}(d) illustrates the impact of varying the number of hidden layers, considering FCN architectures with 1 to 5 layers. Across all configurations, the test loss demonstrates clear power-law scaling with respect to the training dataset size. However, the power-law exponents differ by network depth: shallower networks (\(n_l = 1, 2\)) exhibit smaller exponents, indicating a reduced capacity to improve performance as dataset size increases. In contrast, deeper networks (\(n_l = 3, 4, 5\)) show nearly saturated exponents with only marginal gains from added layers, suggesting diminishing returns in scaling benefits beyond a certain depth. This behavior aligns with findings in the literature, where the scaling exponent increases with model complexity but tends to saturate as networks become deeper, reflecting intrinsic limitations in optimization and model expressiveness for practical dataset sizes~\cite{meir2020powerlaw,
bahri2024explaining}.

\begin{figure}[t!]
    \centering
    \includegraphics[width=0.4\textwidth]{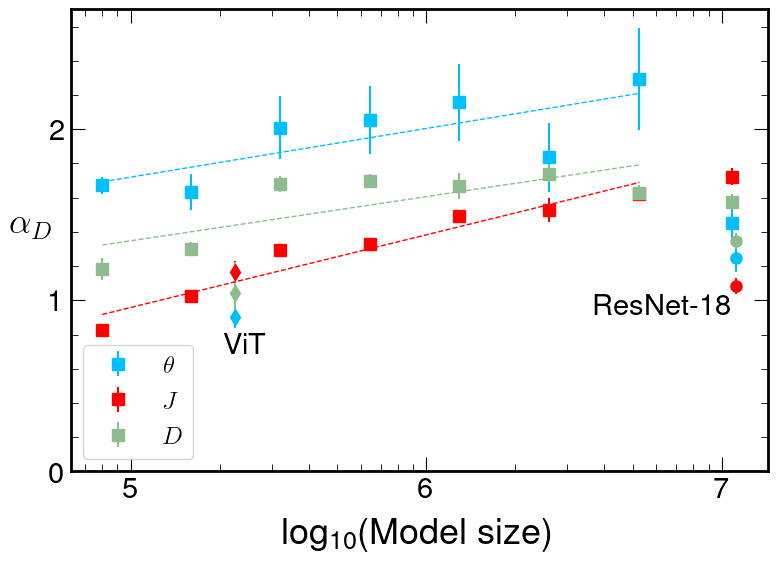}
    \caption{
    \textbf{Power-law scaling exponents \(\alpha_D\) as a function of FCN model size.} Each marker displays the fitted parameter of \(\alpha_D\) from the MSE test loss data in Fig.~\ref{fig4}(a)–(c) for \(\theta\), \(J\), or \(D\), respectively. The FCN model size \(N_M\) ranges from approximately \(8 \times 10^{4}\) to \(10^{7}\) parameters, corresponding to FCNs with \(n_n\) from 4 to 512 and \(n_l=3\). Dashed lines represent logarithmic fits of the form \(\alpha_D = a_D \log_{10}(N_M/10^6) + b_D\),  where \(a_D\) and \(b_D\) are fitting parameters. The data points for the largest value of \(N_M\) are omitted from the fit. Circle and diamond markers indicate the \(\alpha_D\) values for ResNet-18 and ViT, respectively.
    }
    \label{fig5}
\end{figure}

Figure~\ref{fig5} shows the power-law scaling exponents \(\alpha_D\), derived from the test loss data in Fig.~\ref{fig4} by fitting to \(\epsilon \sim N_D^{-\alpha_D}\). The values of \(\alpha_D\) vary considerably with model size: ranging from 1.4 to 2.3 for \(\theta\), 0.8 to 1.8 for \(J\), and 1.2 to 1.8 for \(D\). Notably, the exponents tend to increase with model size. The logarithmic fit for the dataset size exponent, \(\alpha_D = a_D \log_{10}(N_M/10^6) + b_D\), matches the data well for \(J\), while it is less pronounced for \(D\) and \(\theta\); we emphasize that here the subscript \(D\) is associated with the dataset size not the magnetic parameter \(D\). The fitted parameters are: \(a_D = 0.28 \pm 0.11, b_D = 2.00 \pm 0.07\) for \(\theta\); \(a_D = 0.42 \pm 0.05, b_D = 1.38 \pm 0.03\) for \(J\); and \(a_D = 0.26 \pm 0.10, b_D = 1.61 \pm 0.06\) for \(D\). Large error bars are observed primarily for \(\theta\), making conclusions for this parameter less definitive. Additionally, the exponents for \(\theta\) and \(D\) decline at the largest dataset sizes, deviating from the overall increasing trend. This likely results from a strong dependence of the test loss on the training scheme, suggesting potential for further network optimization. Therefore, more refined analysis may be needed to accurately characterize behavior in this large dataset regime. For comparison, the exponents for ResNet-18 and ViT, derived from the test loss data in Fig.~\ref{fig1}, are also displayed in Fig.~\ref{fig5}. These do not follow the same logarithmic pattern as inferred from FCN, despite involving the same task and their comparable model sizes.

We observed that the average of MSE test loss, computed using either the arithmetic mean or median, exhibits scaling behaviors similar to the geometric mean: in both cases, the test loss decreases monotonically with increasing dataset size, with more pronounced reductions for larger models. Their scaling exponents show quantitative agreement with those from the geometric mean. This consistency can be attributed to the convergence of the geometric mean, arithmetic mean, and median for sufficiently large models [Figs.~\ref{fig3}(b) and (d)]. Consequently, these measures can be used interchangeably for large models. However, adopting the geometric mean provides a more robust and consistent framework for evaluating model performance across all network sizes. Additionally, we have examined other activation functions, including ReLU, ELU, PReLU, and Swish. All results demonstrate power-law scaling behaviors consistent with the GeLu case. A key difference lies in the values of the scaling exponents, which deviate by at most 3\% for \(J\) and 28\% for \(D\). Further details are available in the Supplementary Information. 

In summary, we find that, for all network architectures, FCN, ResNet-18 and ViT, the test loss exhibits well-defined power-law scaling with dataset size \(N_D\). Additionally, for FCN, increasing model parameters generally leads to higher scaling exponents.

\subsection{Scaling laws with the model size}

\begin{figure}[t!]
    \centering
    \includegraphics[width=0.75\textwidth]{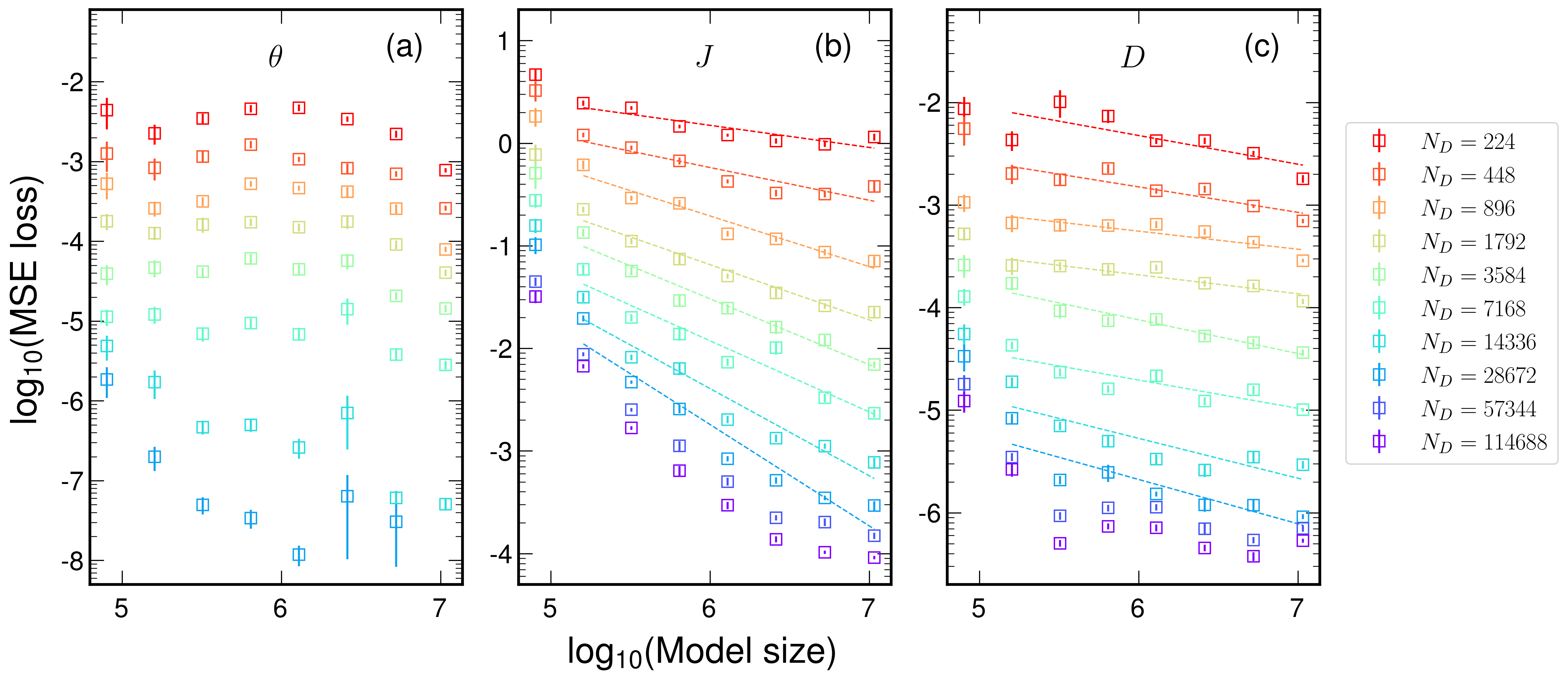}
    
    \caption{
    \textbf{Geometric mean of MSE test loss as a function of FCN model size.} Panels (a)–(c) show results for the regression of \(\theta\), \(J\), and \(D\), respectively. Each marker represents different training dataset sizes ranging from 224 to 114688. In panels (b)–(c), dashed lines are power-law fits of the form \(\epsilon \sim N_M^{-\alpha_M}\), with the scaling exponent \(\alpha_M\). All data are plotted on a log-log scale, with the training dataset size ranging from 224 to 114688. 
    }    
    \label{fig6}
\end{figure}

Figure~\ref{fig6} shows the MSE test loss data from Fig.~\ref{fig4}(a)–(c), plotted as a function of FCN model size at a fixed training dataset size.
The test loss for \(\theta\) generally decreases with increasing model size, but exhibits nonmonotonic, irregular patterns that do not follow a power-law scaling [Fig.~\ref{fig6}(a)]. In contrast, the test loss for \(J\) shows a more consistent, monotonic decreasing trend with increasing model size, spanning several orders of magnitude [Fig.~\ref{fig6}(b)]. Notably, this reduction becomes more pronounced with larger dataset sizes, resulting in a more uniform decline toward the largest model size considered. Conversely, for smaller dataset sizes, the test loss tends to plateau at higher model sizes and, in some cases, shows a slight upturn—for example, in the curves for \(N_D=224, 448\). A similar decreasing trend, with substantial reduction at larger dataset sizes, is also observed for parameter \(D\) [Fig.~\ref{fig6}(c)]. However, in this case, the decrease is more modest and accompanied by notable fluctuations at small and large dataset sizes. Only intermediate dataset sizes display a monotonic decreasing trend. The fluctuations at the two largest dataset size (\(N_D=57344, 114688\)) are likely due to the test loss approaching the limits of numerical precision approximately given by \(1.9 \times 10^{-7}\). This constraint is imposed by the 32-bit floating-point arithmetic used for the computation of the regressed parameters and the corresponding MSE test loss.

\begin{figure}[t!]
    \centering
    \includegraphics[width=0.4\textwidth]{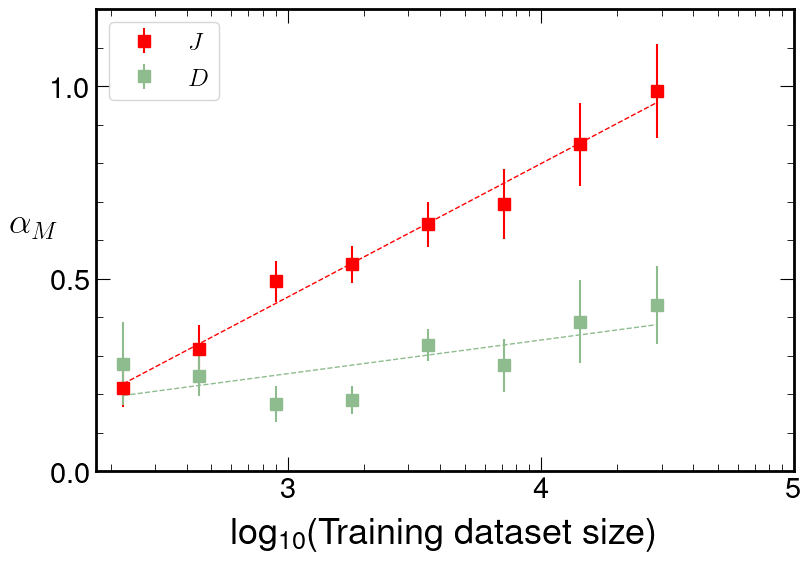}
    
    \caption{
    \textbf{Power-law scaling exponents \(\alpha_M\) as a function of training dataset size.} Each marker displays the fitted parameter of \(\alpha_M\) from the MSE test loss data in Fig.~\ref{fig6}(b)–(c) for \(J\) and \(D\), respectively. The training dataset size \(N_D\) ranges from 224 to 114688. The dashed lines represent logarithmic fits of the form \(\alpha_M = a_M \log_{10}(N_D/10^4) + b_M\), where \(a_M\) and \(b_M\) are fitting parameters.
    }
    \label{fig7}
\end{figure}

Figure~\ref{fig7} shows the power-law scaling exponents \(\alpha_M\), obtained from fitting the test loss data for parameters \(J\) and \(D\) in Fig.~\ref{fig6}(b)–(c) to the relation \(\epsilon \sim N_M^{-\alpha_M}\). The \(\alpha_M\) values for \(J\) increase significantly from \(0.22 \pm 0.05\) to \(0.99 \pm 0.12\) as the dataset size grows. These exponents are well described by a logarithmic fit, \(\alpha_M = a_M \log_{10}(N_D/10^4) + b_M\), with fitted parameters \(a_M = 0.347 \pm 0.018\) and \(b_M = 0.799 \pm 0.017\). In contrast, the \(\alpha_M\) values for \(D\) show only a weak increasing trend with the dataset size within a small range from \(0.18 \pm 0.05\) to \(0.43 \pm 0.10\), without significant variation or clear monotonicity. The fitted parameters in this case are \(a_M = 0.087 \pm 0.035\) and \(b_M = 0.341 \pm 0.032\), with large uncertainties, indicating no consistent increase. The data for the two largest dataset sizes (\(N_D=57344, 114688\)) do not conform well to the power-law fit for both $J$ and $D$ and are omitted from Fig.~\ref{fig7}.

In summary, the test loss for the regression of \(J\) and \(D\) exhibits power-law scaling behavior with respect to model size \(N_M\) at small and intermediate dataset sizes, while the logarithmic increase of \(\alpha_M\) is only evident for \(J\). Conversely, the test loss for the regression of \(\theta\) does not exhibit power-law scaling with respect to \(N_M\), despite decreasing significantly as \(N_M\) increases.

\section{Discussion}

In this study, we demonstrate the presence of a neural scaling law in deep regression through the widespread power-law relationships observed across several orders of magnitude in dataset size, model size, and diverse network architectures—including FCN, ResNet-18, and ViT. The large scaling exponents, reaching up to 2.3 with respect to dataset and model sizes, suggest that the accuracy of deep regression can be significantly improved simply by increasing data resources, or even more substantially by employing large models. The detailed scaling laws presented here offer valuable insights for designing deep regression models with predictable performance based on dataset size and network architecture, thereby mitigating resource-intensive trial-and-error experimentation. These findings are especially beneficial for applications in the physical sciences~\cite{Dunjko2018, Carleo2019, Mehta2019}, such as condensed matter physics \cite{ohtsuki2016deep2d, Carrasquilla2017, PhysRevE.95.062122, ohtsuki2020drawing, Bedolla_2021, Lee2023, doi:10.1126/sciadv.abb0872, kwon2021magnetic, Lee2021, doi:10.1021/acsami.2c12848, Cadez2023MLFractalPhases, Bayo2025MLCondensedMatter, Lee_2024}, high energy and particle physics~\cite{Radovic2018, Albertsson2018, Guest2018} and astrophysics~\cite{Hezaveh2017, Ntampaka2019, Reiman2020}, where data often come in the form of simulated or experimental images, but data resources are frequently limited.

The emergence of substantial scaling exponents in generalization error exceeding two is noteworthy, especially since previous reports typically observe exponents less than one. For example, Hestness et al. found exponents in the range 0.07–0.35 for image classifications and language modeling \cite{hestness2017deeplearningscalingpredictable}; Kaplan et al. reported an exponent of 0.095 for language modeling \cite{kaplan2020scalinglawsneurallanguage}; Henigham et al. observed exponents between 0.037 and 0.24 for image classifications \cite{henighan2020scaling}; Bahri et al. reported exponents ranging from 0.26 to 1.22 for image classifications \cite{bahri2024explaining}; and other studies have shown similar trends \cite{BAILLY2022106504, rosenfeld2020constructive, meir2020powerlaw, mühlenstädt2024dataneed2predicting, Figueroa2012, cho2016dataneededtrainmedical, 8237359, JMLR:v23:20-1111, NEURIPS2022_8c22e5e9, hoffmann2022trainingcomputeoptimallargelanguage, hutter2021learningcurvetheory,Defilippis2023DimensionFree,Paquette2024phases}. The first three studies used larger datasets and model sizes than those in this work, while the last study employed datasets and models of similar scale. This may explain the smaller exponents observed in the initial studies, with some overlap in exponents compared to our results and the last work. However, the absence of large scaling exponents in smaller dataset size regimes in the first three studies \cite{hestness2017deeplearningscalingpredictable, kaplan2020scalinglawsneurallanguage, henighan2020scaling} challenges this explanation, suggesting that other factors (assigned regression tasks and specific scaling analysis methodologies) may also play significant roles. Additionally, network architecture appears to be a key factor; we found that FCNs tend to exhibit larger scaling exponents than ResNet-18 and ViT, a trend also observed in other studies. A detailed comparison and a comprehensive overview of the currently identified scaling laws are provided in the Supplementary Information.

Generalizing the deep regression neural scaling laws established here requires case-by-case analyses to identify power-law relationships across various regression tasks, both within the physical sciences and beyond. This is supported by the observed significant dependence of the scaling exponent on different regression tasks for each magnetic parameter studied. A key step toward real-world applications of these scaling laws is to investigate how imperfections in datasets—such as thermal noise in physical systems—affect the qualitative and quantitative behavior of deep regression scaling laws. An important question is whether high values of the scaling exponents are generic, suggesting that increasing the amount of training data would reliably lead to substantial performance gains. Developing a theoretical framework to explain the emergence of power-law scaling in deep regression—based on the empirical findings presented here, which currently remain largely unexplained—is a critical research direction that can inspire further advancements and deepen understanding in this field.

\section{Methods}

\subsection{Dataset generation}

To generate magnetic domain images for twisted van der Waals magnets, the following Heisenberg spin model for twisted bilayer CrI\textsubscript{3} \cite{Kim2023} was employed:
\begin{equation}
    H = -J\sum_{l=t,b} \sum_{\langle i, j\rangle} \bm{S}_{i}^l \cdot \bm{S}_{j}^l - D \sum_{l=t,b} \sum_{i} \big(\bm{S}_{i}^l\cdot\hat{z}\big)^2 + \sum_{i, j} J_{ij}^\perp \bm{S}_{i}^t \cdot \bm{S}_{j}^b. \label{eq:spinH}
\end{equation}
Here, $\bm{S}_{i}^{l}$ represents the spin operator at $i$ site on the top layer for $l=t$ or the bottom layer for $l=b$. The parameter $J>0$ represents intralayer ferromagnetic (FM) exchange interactions, $D>0$ represents single-ion anisotropy energy, and $J_{ij}^\perp$ represents interlayer exchange interactions. All these magnetic parameters are measured in meV units. The system is constructed on a commensurate moiré superlattice comprising two honeycomb lattice layers rotated relative to each other by a twist angle $\theta$. The twist angle $\theta$ determines the periodicity of $J_{ij}^\perp$, which is approximately given by $L \simeq a/\theta$, where $a$ is the lattice constant of the honeycomb lattice. In total, the spin model is characterized by the four parameters $J$, $D$, $J_{ij}^\perp$, and $\theta$. The spin model is particularly relevant to the material CrI\textsubscript{3} for the parameter values of $J \approx 2$ and $D \approx 0.2$ \cite{PhysRevX.8.041028}.

To obtain diverse magnetic domain images, parameter sets ($\theta$, $J$, $D$) were randomly sampled within the following ranges: $\theta$ from 1.01\textdegree{} to 3.89\textdegree{}, $J$ from 1 to 10, and $D$ from 0.01 to 0.3. The values of $J_{ij}^\perp$ were fixed, as determined by previous ab initio calculations \cite{Kim2023}. Parameter sets corresponding to the FM state were excluded, since the FM state does not exhibit magnetic domains or variation with respect to parameter changes; excluding these states helps prevent potential overfitting \cite{Hawkins2004}. In total, 162,782 parameter sets were used in our analysis. The magnetic ground states for each set were obtained through atomistic spin simulations. From each ground state, two magnetic domain images of size \(100 \times 100\) pixels were generated, each depicting out-of-plane magnetization in either the top or bottom layer. Consequently, our dataset comprises 325,564 paired images that illustrate the morphological variations of magnetic domains in twisted bilayer ferromagnets associated with the parameter changes. Examples of these images are shown in Fig.~\ref{fig:model}(a), where the parameters used are \(J = 1, 2, 3, 4, 5\,\text{meV}\), arranged from the first to fifth columns, and \(\theta = 1, 2, 3^\circ\), arranged from the first to third rows. Further details about the data generation process can be found in the Supplementary Information.

\subsection{Deep learning algorithms}

A test set of 20000 paired images was randomly selected from the total dataset of 162,782 paired images. The remaining data was bootstrapped to sizes of \(N_D = 256, 512, \ldots, 131072\), and each subset was divided into training and validation datasets with an 7/8 and 1/8 split, respectively. Pixel values of all images were normalized and standardized to enhance network performance. Each network was trained using the Adam optimizer for 200 epochs with early stopping patience set to 100. An adjustable learning rate was employed, starting at 0.001 and decreasing by a factor of 0.98 after each epoch. This training scheme was optimized to achieve the best performance, especially for the largest dataset, as detailed in the Supplementary Information. For each network architecture and dataset size, experiments were repeated approximately 20 times for FCN and ResNet, and around 10 times for ViT. All deep learning training and evaluation were conducted using PyTorch~\cite{PyTorch2019} on the Google Colab platform.

\textit{Fully connected network}—Most employed FCNs consist of \( n_l = 3 \) hidden layers, with each hidden layer containing \( n_{n} = 4, 8, \ldots, 512 \) neurons. The input images' pixel values were flattened and concatenated into a single feature vector of length \( n_i = 20{,}000 \). The GELU activation function was used between all layers. The network outputted a single value corresponding to the Hamiltonian parameter via regression. The number of trainable parameters (weights and biases) -- the model size $N_M$, is given by: \(N_M = n_{n} \left[ n_i + (n_l - 1)n_{n} + 1 + n_l \right] + 1 \approx n_i n_{n}\), since \( n_i \gg n_{n}, n_l, 1 \).

\textit{Residual network}—The convolutional input layer of ResNet-18 was customized to consist of two input channels, instead of the standard three, to accommodate two grayscale images of size $100 \times 100$ pixels, each for the top and bottom layer domain structures, respectively. The other hyperparameters of the input layer were kept unchanged. Additionally, the output layer was modified to produce a single scalar value for regression, as opposed to the original 1000-class classification. Consequently, the number of trainable parameters was reduced to 11,173,889 from the original 11,689,512. Due to these modifications, pretrained weights were not used; instead, the network was initialized randomly and trained from scratch.

\textit{Vision transformer}—The vision transformer was customized with a patch size of \(10 \times 10\), resulting in 200 tokens per pair of input images, close to the 196 tokens used in the original ViT. The model utilized multi-head self-attention with 8 attention heads, 4 transformer layers, an embedding dimension of 64, and an MLP size of 256. This configuration comprised a total of 225,601 trainable parameters. The training was conducted for 200 epochs, and we confirmed that further increasing the number of epochs yielded only minor improvements in the test loss. Additionally, L2 and dropout regularization were not employed, as they did not enhance the test loss.

\section*{Acknowledgement}
T.\v{C}. and K.-M.K. were supported by an appointment to the JRG Program at the APCTP through the Science and Technology Promotion Fund and Lottery Fund of the Korean Government. This research was also supported by the Korean Local Governments - Gyeongsangbuk-do Province and Pohang City. \\

\bibliography{ref}

\end{document}